# CUSUM Filter for Brain Segmentation on DSC Perfusion MR Head Scans with Abnormal Brain Anatomy

Svitlana Alkhimova
Department of Biomedical Cybernetics,
National Technical University of Ukraine "Igor Sikorsky Kyiv Polytechnic Institute"
37 Prosp. Peremohy Kyiv 03056 UKRAINE
+380442048574
asnarta@gmail.com

## ABSTRACT
This paper presents a new approach for relatively accurate brain region of interest (ROI) detection from dynamic susceptibility contrast (DSC) perfusion magnetic resonance (MR) images of a human head with abnormal brain anatomy. Such images produce problems for automatic brain segmentation algorithms, and as a result, poor perfusion ROI detection affects both quantitative measurements and visual assessment of perfusion data. In the proposed approach image segmentation is based on CUSUM filter usage that was adapted to be applicable to process DSC perfusion MR images. The result of segmentation is a binary mask of brain ROI that is generated via usage of brain boundary location. Each point of the boundary between the brain and surrounding tissues is detected as a change-point by CUSUM filter. Proposed adopted CUSUM filter operates by accumulating the deviations between the observed and expected intensities of image points at the time of moving on a trajectory. Motion trajectory is created by the iterative change of movement direction inside the background region in order to reach brain region, and vice versa after boundary crossing. Proposed segmentation approach was evaluated with Dice index comparing obtained results to the reference standard. Manually marked brain region pixels (reference standard), as well as visual inspection of detected with CUSUM filter usage brain ROI, were provided by experienced radiologists. The results showed that proposed approach is suitable to be used for brain ROI detection from DSC perfusion MR images of a human head with abnormal brain anatomy and can, therefore, be applied in the DSC perfusion data analysis.

## 1. INTRODUCTION
Nowadays such non-invasive method of perfusion evaluating technique as DSC perfusion MR imaging plays significant role in diagnostic and management of cerebrovascular and oncological diseases [1-3]. This technique provides the recording of signal changes on dynamic series of MR images. Signal changes are caused by the passage of contrast-agent particles through the exam volume that creates local susceptibility inhomogeneity.

Pixel-by-pixel processing of DSC perfusion MR images allows to quantify different hemodynamic parameters and is used to generate color-coded perfusion maps. As a result of DSC perfusion data analysis, physicians can obtain quantitative perfusion values or their visual interpretation (perfusion maps).

Removal of non-target image pixels such as image noise or skull and scalp pixels is an important step in DSC perfusion data analysis. Involving of such pixels data in perfusion analysis leads to the presence of numerous artifacts on perfusion maps and can cause falsely high or falsely low results of perfusion parameters assessment [4, 5]. Thus, software for DSC perfusion data analysis should operate with pre-processing tools for brain ROI detection, i.e. is should maintain brain segmentation on DSC perfusion MR images of a human head with abnormal brain anatomy [4, 6].

## 2. PROBLEM STATEMENT
Most modern software for DSC perfusion data analysis provides more or less automated tools for brain ROI detection on DSC perfusion MR images.

At present, from the clinical practice point of view, automated segmentation is preferable. It can be explained in terms of time-consuming, reproducibility, and removing subjectivity from this procedure. Furthermore, manual segmentation requires from operator prior knowledge on brain structures and their pathologies visual presentation on MR images. Despite the benefits listed above, there are also disadvantages to relying on automated segmentation [5, 7]. In case of processing DSC perfusion MR images of a human head with abnormal brain anatomy, high amount of lesions types (tumor, stroke, necrosis, etc) and their challenging shape or appearance in the brain are the reason of automated segmentation fails, and as a result, inaccurate perfusion analysis output. This leads to the existence of a large number of different automatic algorithms for brain image segmentation [8-10].

In most cases, each of existed algorithms utilizes one of the common principles for segmentation: thresholding [11], clustering [12, 13], pattern recognition [9, 14]. In case of intensity based segmentation algorithms that deal with data thresholding or data clustering, incorrect results are caused by overlapping pixel intensities in lesion regions and regions which are targeted to be excluded from the image (background pixels). In case of pattern recognition, at the present moment, there is a lack of pre-segmented templates and training samples for different shape, density, and location of the lesion for such algorithms applying on images with abnormal brain anatomy. Considering the fact that DSC perfusion data is T2-weighted MR images, such data is more complicated for automated brain segmentation than T1-weighted MR images. It can be explained with fatty tissues presence between brain and skull on T2-weighted MR images. Thus, a lot of automatic algorithms for brain segmentation are focused on T1-weighted MR images. Intensities parameterization [15] provides the possibility to transform T2-weighted image intensity onto a standardized T1-weighted intensity scale. It usage allows finding brain region on T2-weighted MR images with any suitable for T1-weighted MR images segmentation algorithm. However, mentioned here issues with processing abnormal brain anatomy images also existent in algorithms oriented on T1-weighted MR images segmentation. It should also be noted that there are segmentation algorithms for T2-weighted brain MR images segmentation that use benefits from specific image acquisition technique [16] or orient on multimodal image processing [17].

The purpose of this study is to adapt CUSUM filter for brain segmentation purposes on DSC perfusion MR images which can provide a more viable alternative to existing algorithms because of accurate ROI detection in case of abnormal brain anatomy.

The rest of the paper is organized as follows. Section 3 discusses the proposed adaptation of CUSUM filter for brain segmentation purposes on DSC perfusion MR images. Section 4 presents and discusses the segmentation results with adapted CUSUM filter usage. Finally, section 5 concludes the paper.

## 3. MATERIALS AND METHODS

CUSUM filter was applied to solve the issue of tracking the boundary by autonomous vehicles [18], and after that, it was spread to segmentation tasks [19].

Image segmentation that is based on CUSUM filter usage is a process of boundary points detection. In such a case, each boundary point is an output of decision function, which determines whether the image point belongs to one image region (foreground) or to another one (background).

In order to detect boundary points decision function is applied to iteratively accumulated points at the time of moving on a trajectory along the boundary between two image regions. The motion trajectory should start with a point at the target boundary and can be further defined by an iterative change of movement direction in one region in order to reach another one, and vice versa. Basic point-by-point algorithm for iterative creation of motion trajectory can have only three parameters: turning direction value, turning angle value, and step size [20]. The algorithm creates a motion trajectory through operating with these three parameters; as a result, it accumulates image points for further filtering by CUSUM decision function.

Proposed by Page [21] CUSUM filter is a time-weighted statistical control that uses cumulative sums to detect the deviation of each observed value from the expected value and notify about a moment of time when monitored process undergoes a change. In statistical terms, the CUSUM filter is applied to change-point detection. Alarm time $\tau$ when change is detected and the process becomes the out of control is defined with original CUSUM rule as follows:

$$\tau = \min(i > 0 : S_i > H) \quad (1)$$

where H is the decision interval, which acts as a control limit and means that process is out of control; $S_i$ is cumulative sum up to and including $i^{th}$ sample. The cumulative sum $S_i$ is based on the log-likelihood ratio and defined recursively as follows:

$$S_i = \max\left(0, S_{i-1} + \ln\left[\frac{g_1(z_i)}{g_0(z_i)}\right]\right) \quad (2)$$

where $g_0$ and $g_1$ are probability density functions of the monitored process characteristics $z_i$ before and after the change-point respectively. So, change-point is considered to be found at the step when the value of cumulative sum exceeds decision interval h. At the time of continuous process monitoring after change-point detection happen cumulative sum value is reset to zero and the surveillance continues.

The above formulas can be used only for changes detection in a positive direction; they define so-called one-sided upper CUSUM filter. When changes detection should be done in a negative direction, the max operation in (2) should be replaced with min operation, and this time change-point is considered to be found at the step when the value of cumulative sum in (1) is below the negative value of the decision interval h. Such modified formulas define so-called one-sided lower CUSUM filter.

In case of image processing, each monitored process characteristics $z_i$ is image intensity $I_i$ at point i on a motion trajectory. Thus CUSUM filter should detect a change-point in a moment of crossing from one image region into another. Considering that actual data distribution of such monitored process is unknown, probability density functions $g_0$ and $g_1$ are unknown as well. Therefore cumulative sum calculation can be done by replacing the log-likelihood ratio in (2) with an appropriate score function, which is sensitive to expected changes [20].

Due to the fact that in common case mean of intensities of image points varies considerably for different regions at places of their boundary location, it was assumed that $g_0$ and $g_1$ can be generated by Gaussians [19, 20]. According to this assumption cumulative sum calculation for two-sided CUSUM filter can be defined recursively as follows:

$$\begin{cases} S_i^+ = \max(0, S_{i-1}^+ + I_i - (\mu_0 + K)) \\ S_i^- = \min(0, S_{i-1}^- + I_i - (\mu_0 - K)) \end{cases} \quad (3)$$

where $I_i$ is image intensity at point i; $\mu_0$ is the target mean (mean of the previous observations of intensities of image points in the target region); K is reference value; $S^+$ and $S^-$ are cumulative sums for upper-sided (signals if $S^+ > H$) and lower-sided (signals if $S^- < -H$) CUSUM filters respectively. The starting values of $S^+$ and $S^-$ are equal to 0. At some unknown time the mean $\mu_1$ of the observations of intensities of image points undergoes a shift of a fixed size $\Delta$, i.e. $\mu_1 = \mu_0 + \Delta$, and this value is considered to be out of control. Often $\Delta$ is measured in multiples of $\sigma$, that is $\Delta = \delta\sigma$. K and H are two important parameters for CUSUM filter. For CUSUM filters of normal data K is generally set to $\Delta/2$ and H is defined as $5\sigma$, then the process is considered to be out of control [22].

In case of brain ROI detection on DSC perfusion MR images, segmentation fails with usage of CUSUM filter according to (3). The normal distribution can fail due to one or more of the following: the actual data distribution may be heavier-tailed than normal, so it can cause occurrence of far-out intensity values; true intensity values may be corrupted by motion artifacts or noise and such value is a far-out value; it may be an isolated special cause that affects an individual case.

Segmentation results of DSC perfusion MR image with usage of CUSUM filter according to (3) are shown in Figure 1, where three different zoomed-in samples show brain boundary detection failure.

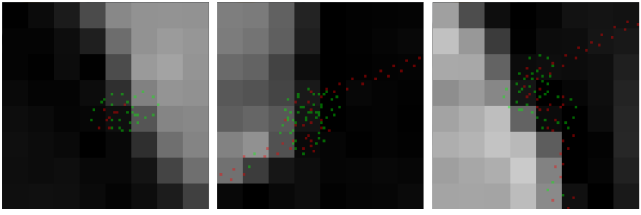

**Figure 1. Failed boundary detection with CUSUM filter that is defined according to the assumption of a normal distribution (zoomed-in samples of DSC perfusion MR image; found boundary points are marked with red).**

From Figure 1, it can be deduced that used CUSUM filter according (3) is not capable of handling boundary crossing and motion trajectory sticks around the same group of pixels until the value of decision interval is corrupted due to the wrong accumulated intensity values.

In order to adapt CUSUM filter to DSC perfusion MR images segmentation, the current study proposes to calculate cumulative sum for two-sided CUSUM filter recursively as follows:

$$\begin{cases} S_i^+ = \max(0, S_{i-1}^+ + I_i - \mu_1) \\ S_i^- = \min(0, S_{i-1}^- + I_i - \mu_1) \end{cases} \quad (4)$$

where $I_i$ is image intensity at point i; $\mu_1$ is the mean of the previous observations of intensities of image points in current region; $S^+$ and $S^-$ are cumulative sums for upper-sided and lower-sided CUSUM filter respectively. Change-point is considered to be found at the step when $S^+ > \mu_0 - \mu_1$ for upper-sided CUSUM filter and $S^- < -(\mu_0 - \mu_1)$ for lower-sided CUSUM filter.

Segmentation results of the same as on Figure 1 DSC perfusion MR image with usage of adapted CUSUM filter according to (4) are shown on Figure 2.

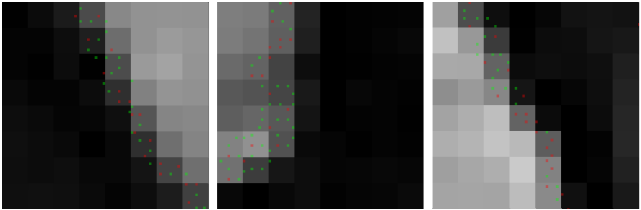

**Figure 2. Boundary detection with adapted CUSUM filter (the same as on Figure 1 zoomed-in samples of DSC perfusion MR image; found boundary points are marked with red).**

Both on Figure 1 and Figure 2 CUSUM filters were applied to the same 4th time-point image from dynamic series of DSC perfusion MR images, on which signal intensity is reached a steady state.

## 4. RESULTS AND DISCUSSION

The segmentation algorithm with proposed adopted CUSUM filter usage was applied to detect brain ROI on DSC perfusion MR images from 6 clinical cases. MR head scans were acquired on a 3.0 T clinical MR scanner (Achieva, Philips Healthcare, Best, the Netherlands) from 6 patients with cerebrovascular disease. Scan parameters were: repetition time = 1500 ms, echo time = 30 ms, flip angle = 90°, field of view = 23 x 23 cm, image matrix = 128 x 128, slice thickness = 5 mm, and gap = 1 mm. 17 slices were scanned with 40 dynamic images for each slice. Contrast medium (Gadovist, Bayer Schering Pharma AG, Germany) with a dosage of 0.1 mmol/kg body weight was injected at a rate of 5 mL/sec, followed by a 10-mL bolus of normal saline also at 5 mL/sec. All images were collected in 12-bit DICOM (Digital Imaging and Communication in Medicine) format.

The segmentation algorithm with adopted CUSUM filter usage was implemented in C# and was applied to the images without any preprocessing such as noise reduction, motion correction or intensity nonuniformity correction. The algorithm was developed as independent of processed image resolution.

Segmentation process was performed using the 4th time-point image, on which signal intensity is reached a steady state, for every spatial position of the input image series. Segmentation output was a binary mask that has unity values for brain ROI pixels. The binary mask was generated based on brain boundary, which was detected with adopted CUSUM filter. The initial point of motion trajectory was set manually at the target boundary between the brain and surrounding tissues.

The average computation time of brain boundary detection and binary mask generation was about 0.2s per image (Intel Core i5-460M 2.53 GHz, single threaded), which is quite fast.

Obtained segmentation results were compared with a reference standard, which is the manually marked brain ROI by an experienced radiologist and confirmed by a second radiologist. 10 sample images were selected from each of the clinical cases as the image set for the validation experiments.

Dice index was calculated for all 60 sample images in order to estimate spatial overlap between two ROIs on each image as follows:

$$\text{Dice} = 2|ROI_S \cap ROI_R|/(|ROI_S| + |ROI_R|) \quad (5)$$

by considering detected with CUSUM filter usage brain ROI as $ROI_S$ and provided by experienced radiologists brain ROI as $ROI_R$. Dice index values can range between 0 (no overlap) and 1 (perfect agreement). In the current study it reached 0.9488 ± 0.0358. Hence, it can be considered as a high agreement between the segmentation results and reference standard.

An example of segmentation result for the selected sample image with the average similarity to the reference standard is shown on Figure 3.

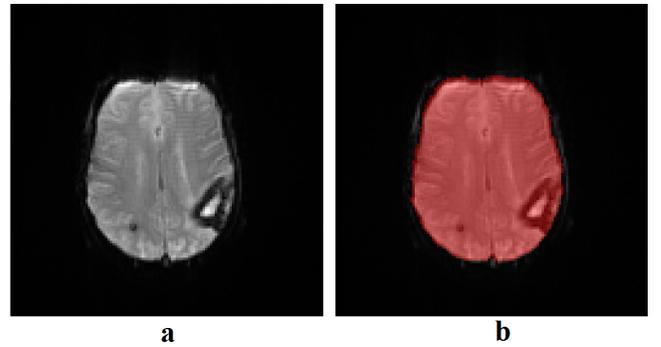

a          b

**Figure 3. Results of brain segmentation with adapted CUSUM filter usage on the selected sample image with the average similarity to the reference standard: a - original image; b - detected brain ROI on a (marked with red).**

The segmentation results also were compared with results, which were obtained with a user-defined threshold method. The comparison was done on the same image set that was selected for the validation experiments. User-defined thresholding is a state-of-the-art method for perfusion ROI detection that is used in brain image processing system for clinical use. More details on optimal threshold selection and obtained results for user-defined threshold method can be found in a previous study [5]. Developed in the current study segmentation algorithm showed superior performance compared with the user-defined threshold method that reached only to 0.7829 ± 0.0866 for evaluating with Dice index metric.

Extracted brain ROI pixels from shown on Figure 3 sample image that were obtained with binary masks of the proposed and user-defined threshold method of segmentation are shown on Figure 4.

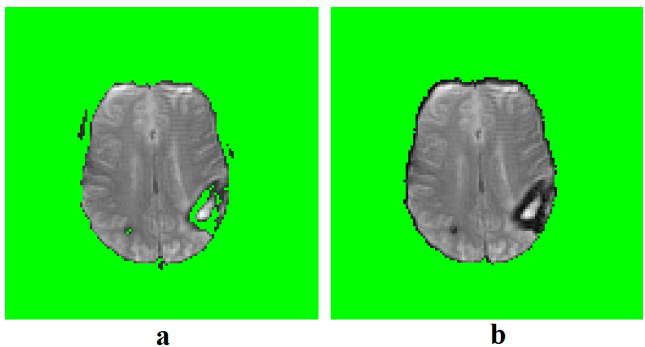

**Figure 4. Extracted brain ROI pixels from shown on Figure 3 sample image: a - by user-defined threshold method; b – by developed in current study segmentation algorithm.**

## 5. CONCLUSION AND FUTURE WORK

New segmentation approach for brain ROI detection on DSC perfusion MR images of a human head with abnormal brain anatomy was proposed in this study. It produces binary mask based on brain boundary, which points are detected as a change-points by adopted CUSUM filter. In order to solve this task it operates by accumulating the deviations between the observed and expected intensities of image points at the time of moving on a trajectory along the boundary between two image regions.

The segmentation algorithm with proposed adopted CUSUM filter usage was developed and tested on 60 DSC perfusion MR images from 6 different patients with cerebrovascular disease. It was achieved good results by evaluating the algorithm with Dice index that reached 0.9488 ± 0.0358 in case of developed algorithm compared to only 0.7829 ± 0.0866 in case of user-defined threshold method. All detected brain ROIs were also evaluated via visual inspection by two experienced radiologists and considered as good enough for clinical use. The computation time was about 0.2s per image of 128x128 pixels in size, which is quite fast.

Now, the segmentation approach with CUSUM filter usage should be tested on more images of different kind and size of lesions in order to better validation. A possible further extension for this study also could be investigating various motion trajectories shapes and different estimators of the CUSUM filter. In addition, it would be important to develop an accurate automatic search for initial boundary point detection to provide fully automated segmentation mode without applying any user interaction. Finally, to provide more accurate perfusion ROI detection results it would be also very important to extract ventricular CSF within obtained in the current study binary mask for brain ROI.